\documentclass[lettersize,journal]{IEEEtran}
\usepackage{amsmath,amsfonts} 
\usepackage{array}

\usepackage{textcomp}
\usepackage{stfloats}
\usepackage{url}
\usepackage{verbatim}
\usepackage{graphicx}
\usepackage{cite}
\usepackage{pifont}
\usepackage{booktabs} 
\usepackage{bm}
\usepackage[T1]{fontenc} 
\usepackage[linesnumbered,ruled,vlined]{algorithm2e}
\usepackage{hyperref}
\usepackage{subcaption}
\usepackage{orcidlink}
\usepackage{makecell}
\usepackage{multirow}

\allowdisplaybreaks

\usepackage{algorithmic}
\usepackage{amsthm}

\newtheorem{remark}{Remark}

\begin{document}

\captionsetup[figure]{labelformat=simple, labelsep=period} 

\captionsetup[table]{name={TABLE},labelsep=space} 

\title{Mitigating Estimation Bias with Representation Learning in TD Error-Driven Regularization}

\author{Haohui Chen~\orcidlink{0000-0001-9660-0948}, Zhiyong Chen~\orcidlink{0000-0002-2033-4249}, Aoxiang Liu~\orcidlink{0009-0000-8047-1192}, and Wentuo Fang~\orcidlink{0000-0002-0374-3534} 

\thanks{Haohui Chen, Aoxiang Liu, and Wentuo Fang are with the School of Automation, Central South University, Changsha 410083, China (e-mail: \href{mailto:haohuichen@csu.edu.cn}{haohuichen@csu.edu.cn}; \href{mailto:ordisliu@csu.edu.cn}{ordisliu@csu.edu.cn}; \href{mailto:wentuo.fang@outlook.com}{wentuo.fang@outlook.com}).}
\thanks{Zhiyong Chen is with the School of Engineering, University of Newcastle, Callaghan, NSW 2308, Australia (e-mail: \href{mailto:zhiyong.chen@newcastle.edu.au}{zhiyong.chen@newcastle.edu.au}).}
\thanks{Corresponding author: Zhiyong Chen.}}

\maketitle

\begin{abstract}
Deterministic policy gradient algorithms for continuous control suffer from value estimation biases that degrade performance. While double critics reduce such biases, the exploration potential of double actors remains underexplored. Building on temporal-difference error-driven regularization (TDDR), a double actor–critic framework, this work introduces enhanced methods to achieve flexible bias control and stronger representation learning. We propose three convex combination strategies, symmetric and asymmetric, that balance pessimistic estimates to mitigate overestimation and optimistic exploration via double actors to alleviate underestimation. A single hyperparameter governs this mechanism, enabling tunable control across the bias spectrum. To further improve performance, we integrate augmented state and action representations into the actor and critic networks. Extensive experiments show that our approach consistently outperforms benchmarks, demonstrating the value of tunable bias and revealing that both overestimation and underestimation can be exploited differently depending on the environment.

\end{abstract}


\section{Introduction}

\IEEEPARstart{A}{ctor}–critic (AC) methods are widely used in reinforcement learning (RL) \cite{Hu2024SurveyPaper}. In the AC framework, accurate value estimation plays a critical role in guiding both policy evaluation and policy improvement \cite{lyskawaACERACEfficientReinforcement2024}, \cite{omura2024symmetric}. Q-learning constructs the target value of the Q-function through a maximization operation; however, this operation often introduces overestimation errors \cite{guo2024acwithsynthesisloss}. Such overestimation makes it difficult for agents to acquire effective policies, and poor policies in turn yield unreliable and biased value estimates \cite{Li2025PrimacyBias}.
 
Deep deterministic policy gradient (DDPG) \cite{lillicrap2019continuouscontroldeepreinforcement} extends Q-learning to continuous action spaces and employs the deterministic policy gradient \cite{silver2014deterministic} for policy updates. However, DDPG also suffers from overestimation due to the maximization operation inherited from Q-learning. To address this, Fujimoto et al. \cite{fujimoto2018addressing}, inspired by double Q-learning \cite{hasselt2016deep}, proposed clipped double Q-learning (CDQ), which mitigates overestimation and achieves improved performance over DDPG. Nevertheless, twin delayed deep deterministic policy gradient (TD3), which incorporates CDQ, can suffer from underestimation bias \cite{cheng2024dualparellelpolicy}. Indeed, double Q-learning itself is inherently prone to underestimation bias \cite{ren2021estimation}.

Overestimation and underestimation biases arise in temporal-difference (TD) learning due to function approximation error and bootstrapping. When function approximators, particularly deep neural networks, are combined with the $\max$ or $\min$ operator in TD target computation, they can induce systematic biases in Q-values. During training, these operators amplify errors by prioritizing actions whose Q-values are already overestimated or underestimated, thereby reinforcing the bias relative to the true action values. Moreover, deep neural networks themselves may contribute additional bias through function approximation errors. Since TD learning relies on bootstrapping, using estimates of the next state–action pair to update current estimates, any bias in the next-step Q-value is propagated backward through the TD target, causing the current estimate to inherit the same overestimation or underestimation.

Building on the double actor–critic (DAC) framework, several advanced algorithms have been proposed, including double actors regularized critics (DARC) \cite{lyu2022efficient}, softmax deep double deterministic policy gradients (SD3) \cite{pan2020softmax}, generalized-activated deep double deterministic policy gradients (GD3) \cite{lyu2023value}, and temporal-difference error-driven regularization (TDDR) \cite{chen2025doubleactorcritic}. These methods introduce various regularization techniques to improve value estimation and consistently demonstrate superior performance. In particular, DARC, SD3, and GD3 specifically address the underestimation bias inherent in TD3.

TDDR \cite{chen2025doubleactorcritic} improves performance over TD3 without introducing additional hyperparameters. Its convergence to the optimal value has been analyzed under both random and simultaneous updating schemes. However, TDDR still exhibits notable limitations. A clear performance gap remains compared with state-of-the-art (SOTA) algorithms that rely on carefully tuned hyperparameters, and extensive experiments show that TDDR does not sufficiently mitigate underestimation bias. These shortcomings motivate this work, which seeks to close the performance gap while addressing both overestimation and underestimation biases.

To this end, we propose an enhanced TDDR framework with two key refinements. First, we introduce a flexible convex combination mechanism that incorporates both symmetric and asymmetric strategies, using pessimistic estimates to mitigate overestimation bias and double actors for optimistic exploration to alleviate underestimation bias. A single hyperparameter governs this mechanism, providing predictable control across the spectrum from overestimation to underestimation. Second, recognizing that peak performance also critically depends on input representation quality \cite{Wang2025FeatureExtractionRL}, we integrate a representation learning module that generates augmented state and action features, which are then fed into the actor and critic networks.
 
The first refinement introduces three core algorithms with different combinations of actors and CDQ: double-actor double-CDQ (DADC), double-actor single-CDQ (DASC), and single-actor single-CDQ (SASC). Each employs a distinct strategy for integrating pessimistic estimates with optimistic exploration. Applying the second refinement to these core algorithms produces their enhanced counterparts, denoted as DADC-R, DASC-R, and SASC-R, respectively. The primary contributions of this work are summarized as follows.

1. We overcome the limitation of TDDR in mitigating underestimation by introducing a TD error–driven convex combination mechanism. Governed by a single hyperparameter, this mechanism calibrates the trade-off between pessimistic and optimistic estimates, enabling predictable and tunable bias control.

2. We design a novel input architecture that incorporates a dynamics-based representation learning module. The key innovation lies in strategically leveraging the learned representations within the actor and critic networks to enhance performance and improve learning stability.

3. We conduct extensive experiments to validate our approach. The results (i) confirm that the convex combination mechanism enables controllable bias adjustment, (ii) show that overestimation and underestimation each offer distinct benefits across different environments, and (iii) demonstrate that, when combined with representation learning, our algorithms achieve competitive results against benchmarks and several SOTA methods.

The remainder of this paper is organized as follows. Section~\ref{sec:Related Work} reviews related work. Section~\ref{sec:Preliminaries} introduces the preliminaries and outlines the motivation for our study. Section~\ref{sec:Proposed Method} details the two proposed refinements. Section~\ref{sec:Experiments} presents experimental results on MuJoCo and Box2D continuous-control tasks and discusses performance relative to benchmark algorithms. Finally, Section~\ref{sec:Conclusion} concludes the paper and highlights potential directions for future research.

\section{Related Work}
\label{sec:Related Work}

\subsection{Overestimation and Underestimation Biases}

Recently, several improved methods have been proposed to reduce overestimation and underestimation biases. DARC \cite{lyu2022efficient} constructs its TD target through a convex combination of the maximum and minimum Q-values derived from double actors, effectively mitigating underestimation. SD3 \cite{pan2020softmax} addresses underestimation by employing a softmax operator to compute the expected Q-value of the next state across a distribution of actions, assigning exponentially greater weight to higher Q-values. GD3 \cite{lyu2023value} counters underestimation by incorporating a generalized activation function and an explicit, learnable positive bias term into its value estimation process. Although effective, these methods all rely on static, predefined regularization strategies. In contrast, our framework uses the TD error as an intrinsic, real-time signal to dynamically modulate the contribution of each Q-function. This allows our method to adaptively weigh the estimates from the DAC framework based on their current stability, rather than relying on fixed strategies.

Lee et al. \cite{lee2023spqr} explored ensemble methods in Q-learning to exploit the diversity of multiple Q-functions. In maxmin Q-learning \cite{lan2021maxminqlearningcontrollingestimation}, target values are constructed by combining multiple Q-functions to reduce both bias and variance. Compared to using multiple Q-functions alone \cite{chen2021randomizedensembleddoubleqlearning}, combining multiple Q-functions with multiple actors \cite{li2023keep} provides not only diverse Q-value estimates but also a richer set of action choices for the agent. Ensemble strategies can improve the stability and accuracy of Q-values by aggregating outputs from multiple independent networks. However, they typically require maintaining and training a large number of networks. Moreover, when all networks are weighted equally, ensembles may fail to fully exploit high-quality estimates, and without carefully leveraging the ensemble architecture, performance gains may be marginal or even fall below those of baseline methods. Although the DAC framework can be viewed as a basic ensemble scheme, DADC, DASC, and SASC do not rely on naive averaging. Relative to multiple actors, double actors can further enhance exploration by following the policy that yields higher returns \cite{lyu2022efficient}. Similarly, relative to multiple critics, double critics avoid the problem of excessive Q-values, which complicates action selection for agents.

Pessimistic underexploration was introduced by Ciosek et al. \cite{ciosek2019better} to describe the underexploration and underestimation caused by the lower-bound approximation of CDQ. Li et al. \cite{liImprovingExplorationActor2024} proposed the weakly pessimistic value estimation and optimistic policy optimization (WPVOP) algorithm, which employs a pessimistic estimate to adjust the TD target. Cetin et al. \cite{cetinLearningPessimismReinforcement2023} developed the generalized pessimism learning (GPL) algorithm, which integrates dual TD targets to counteract the bias from pessimistic estimation, thereby improving value estimation while reducing computational overhead. However, pessimistic estimation may restrict exploration of unseen actions and states. Double actors can alleviate this issue by enhancing exploration and preventing policies from becoming trapped in local optima, thereby promoting the discovery of potentially better actions \cite{lyu2022efficient}. To address the pessimistic underexploration induced by CDQ, DADC, DASC, and SASC systematically integrate pessimistic estimation with optimistic exploration, suppressing estimation bias while encouraging more effective exploration.

\subsection{Representation Learning}

In addition to mitigating estimation bias, enhancing the quality of input representations is fundamental to improving RL performance \cite{Zou2025goalRL}. This challenge is especially pronounced in vision-based RL, where agents must extract meaningful features from high-dimensional pixel observations \cite{Wang2025VisualSafeControl}. Two primary strategies have emerged to address this issue: employing data augmentation as an implicit form of regularization, and learning representations through auxiliary tasks \cite{Poudel2024RegularizedContrastiveRepresentation}.

Data augmentation has proven effective for representation learning, as demonstrated by DrQ \cite{yarats2021image} and RAD \cite{RAD2020Laskin}, where simple random image transformations such as cropping and translation regularize both the value function and the policy, leading to significant performance gains. DrQ-v2 \cite{yarats2021mastering} further demonstrates that data augmentation alone can yield high sample efficiency in pixel-based RL.

In addition to augmentation, integrating auxiliary self-supervised learning (SSL) objectives with RL can further enhance performance. CURL \cite{curl2020Laskin} incorporates a contrastive learning objective into SAC \cite{haarnoja2018softSAC}, where observations from the replay buffer are augmented to form query–key pairs that are then encoded and compared. However, the effectiveness of jointly training RL and SSL is not always guaranteed. Li et al. \cite{nips2022li} showed that simply adding an SSL loss does not consistently outperform well-tuned augmentation strategies and may even destabilize learning. This suggests that the core challenge lies not only in selecting appropriate auxiliary tasks but also in ensuring stable and coordinated integration between representation learning and RL \cite{Jae2025DataEfficientRL}.

To address this, TD7 \cite{Fujimoto2023TD7} introduced state–action representation learning (SALE), which aims to learn embeddings that capture both the structure of the observation space and the transition dynamics of the environment. SALE employs two encoders, one for states and another for state–action pairs, and optimizes a dynamics prediction loss defined as the mean squared error between the current state–action representation and the representation of the next state. The critic network takes as input the state, action, state representation, and state–action representation, while the actor network receives the state along with its representation. In addition, SALE normalizes state and action inputs to both networks using AvgL1Norm, a normalization layer that divides each input vector by its average absolute value across dimensions.

Drawing inspiration from SALE, we adopt a similar training objective for the two encoders but introduce a fundamentally different strategy for leveraging the learned representations, distinguishing our approach from TD7. For the critic networks, we construct an augmented state by concatenating the raw state with its representation and treat the state–action representation as the augmented action. Thus, the critic networks take the augmented state, augmented action, and raw action as inputs. For the actor networks, only the augmented state is used as input.

\section{Preliminaries}
\label{sec:Preliminaries}

RL can be formulated as a Markov decision process (MDP), which consists of five tuples, denoted as $\langle\mathcal{S},\mathcal{A},\mathcal{P},R(s,a),\gamma\rangle$, where the elements represent state space, action space, state transfer probability, reward function, and discount factor. The goal of RL is to find the optimal policy that maximizes cumulative rewards. We briefly describe several relevant reinforcement learning algorithms, including DDPG, TD3, DARC, SD3, GD3 and TDDR, which serve as benchmarks.

\subsection{Baseline Algorithms}

DDPG extends Q-learning to continuous domains by employing deterministic policies to maximize the Q-function, thereby approximating the max operator \cite{lillicrap2019continuouscontroldeepreinforcement}. The policy is updated using the deterministic policy gradient, defined as:
\begin{align}\label{deqn_ex1a}
  \nabla_{\phi}J(\phi) = N^{-1}\sum_{s}(\nabla_a Q_{\theta}(s,a)|_{a=\pi_{\phi}(s)}\nabla_{\phi}\pi_{\phi}(s)),
\end{align}
where $J(\phi)$ is the objective function, typically the expected return we aim to maximize, and $\nabla_a Q_{\theta}(s,a) \big|_{a=\pi_\phi(s)}$ represents the gradient of the action-value function $Q$ with respect to actions, $\nabla_{\phi} \pi_\phi(s)$ represents the gradient of the policy with respect to its parameters $\phi$, and $N$ represents the batch size. The gradient $\nabla_a Q_{\theta}(s,a)$, for updating $\theta$, is usually obtained by minimizing $N^{-1}\sum_s(y-Q_{\theta}(s,a))^2$, where $y$ is the TD target, $y-Q_{\theta}(s,a)$ is the TD error.

DDPG introduces the concept of target networks and utilizes a soft update approach for both policy and value parameters: 
\begin{align}
\theta^{\prime} \leftarrow \tau \theta + (1 - \tau) \theta^{\prime},\;\phi^{\prime} \leftarrow \tau \phi + (1 - \tau) \phi^{\prime}, \label{targetupdate}
\end{align}
where $\theta^{\prime}$ represents the parameters of the critic target network, and $\phi^{\prime}$ represents the parameters of the actor target network. Here, $\tau \ll 1$ is a constant. 

The TD target in DDPG is computed using the target networks as follows: 
\begin{align}\label{TD1}
y = r + \gamma Q_{\theta^{\prime}}(s^{\prime}, \pi_{\phi^\prime}(s^{\prime})+\epsilon),
\end{align}
where $s^{\prime}$ is the next state, and the output of $\pi_{\phi^\prime}(s^{\prime})$ represents the next action, denoted as $a^{\prime}$, incorporating noise $\epsilon$.

A key challenge in DDPG is the overestimation bias arising from the maximization operator. To address this, TD3 introduces CDQ, which is given by: 
\begin{align} \label{TDTD3}
  y = r + \gamma \min_{i=1,2}(Q_{\theta_i^{\prime}}(s^{\prime},a^{\prime})).
   \end{align}
Both DDPG and TD3 use a single actor for the policy improvement. TD3 demonstrates superior value estimation compared to DDPG in MuJoCo \cite{baselines}, but may cause underestimation bias.

The TD target for DDPG \eqref{TD1} can be rewritten as:
\begin{align} \label{TDpsi}
  y = r + \gamma \psi,
\end{align}
where 
\begin{align} 
\psi = Q_{\theta^{\prime}} (s^{\prime},  a^{\prime}).
\label{psiDDPG}
\end{align}
When double critics are used, the value $\psi$ for each critic following DDPG \eqref{TD1} is given as:
\begin{align} \label{TDdoublecritic}
\psi_i =   Q_{\theta_i^{\prime}} (s^{\prime}, a^\prime).
\end{align}
However, this value is not used directly. For example, it is modified in \eqref{TDTD3} to
 \begin{align} \label{hatpsiTD3}
\psi=\min_{i=1,2} Q_{\theta_i^{\prime}}(s^{\prime}, a^\prime).
 \end{align}
It is noted that the same $\psi$, and hence the same TD target $y$, is used for both critics. 
The difference between \eqref{TDdoublecritic} and \eqref{hatpsiTD3} acts as a regularization of the TD error.

For TD3, $\psi$ is defined in \eqref{hatpsiTD3}, where it uses the smaller Q-values from the double critics to compute $\psi$. For DARC, $\psi$ is computed as:
\begin{align} \label{hatpsiDARC}
    \psi = &(1-\nu)\max_{j=1,2}\min_{i=1,2}Q_{\theta^{\prime}_i}(s^{\prime},a_j^\prime) \nonumber \\ 
    & + \nu\min_{j=1,2}\min_{i=1,2}Q_{\theta^{\prime}_i}(s^{\prime},a_j^\prime),
\end{align}
where $\nu$ is the weighting coefficient, $a_1^{\prime}$ is the action generated by $\pi_{\phi_1^{\prime}}(s^{\prime})$, $a_2^{\prime}$ is the action generated by $\pi_{\phi_2^{\prime}}(s^{\prime})$, incorporating noise $\epsilon$. DARC computes $\psi$ by combining the Q-values of the double critics and double actors in a convex combination.

For SD3 and GD3, $\psi$ is constructed as:
\begin{align} \label{hatpsiSD3}
 \psi = \min_{j,i=1,2}Q_{\theta^{\prime}_i}(s^{\prime},a_j^\prime),
\end{align}
which is similar to TD3 but incorporates actions from double actors. 

\subsection{TDDR}

To effectively utilize the DAC framework, a key challenge lies in determining how to combine the estimates. DARC employs a fixed convex combination of the maximum and minimum Q-values derived from the double critics and double actors. SD3 and GD3 use a static minimum operator across four Q-values. In contrast, TDDR introduces a dynamic, TD error–driven selection process to construct the TD target. Specifically, it leverages the target network $Q_{\theta^{\prime}}$ to evaluate both the next-state actions $(s^{\prime}, a_i^{\prime})$ and the current-state action $(s, a)$, and uses the magnitude of the resulting TD errors, $|\delta_i|$, to guide TD target construction, thereby enhancing the stability of estimates from the actor target networks $\pi_{\phi_1^{\prime}}$ and $\pi_{\phi_2^{\prime}}$. 

The process begins with DA-CDQ:
\begin{align*}
  Q_{\theta_i}(s^{\prime},a_i^{\prime}) = \min_{j=1,2}(Q_{\theta_j^{\prime}}(s^{\prime},a_i^{\prime})),
    \end{align*}
for $i=1,2$. 
Next, to gauge the stability of each actor, TDDR calculates their respective TD errors. The TD error of the critic targets $Q_{\theta_i^{\prime}}$ as follows:
\begin{align} \label{deltai}
  \delta_{i=1,2} = r + \gamma \min_{j=1,2}(Q_{\theta_j^{\prime}}(s^{\prime},a_i^{\prime}))-\min_{j=1,2}(Q_{\theta_j^{\prime}}(s,a)).
\end{align}
It is important to note that these TD errors in \eqref{deltai} are not for the critics $Q_{\theta_i}$ nor are they used to update $\theta_i$ directly. However, these TD errors drive the calculation of the regularization of the TD errors of $Q_{\theta_i}$. TDDR then constructs $\psi$ by selecting the actor target network that exhibits a smaller absolute TD error:
\begin{equation} \label{psiTDDR}
\psi= \begin{aligned}
  \begin{cases}
\min_{i=1,2}(Q_{\theta_i^{\prime}}(s^{\prime},a_1^{\prime})),~\text{if}~|\delta_1|\leq|\delta_2|\\
\min_{i=1,2}(Q_{\theta_i^{\prime}}(s^{\prime},a_2^{\prime})),~\text{if}~|\delta_1|> |\delta_2|
  \end{cases}.
  \end{aligned}
  \end{equation}
This TD error-driven selection serves as a stabilizing mechanism by utilizing the smaller values.

\subsection{DAC for Better Exploration}

The $\min$ operator in CDQ produces a systematically pessimistic lower bound of the true value estimate. A deterministic policy gradient then selects the action that maximizes this pessimistic lower bound. The combination of a greedy update mechanism with an inaccurate pessimistic estimate leads to pessimistic underexploration \cite{lyu2022efficient}. Double actors can effectively enhance the agent’s exploration ability, helping it maintain a balance between exploration and value estimation so that it benefits from both pessimistic estimation and optimistic exploration \cite{pan2020softmax}.

Double actors promote policy diversity, enabling the agent to escape local optima and approach the global optimum. In contrast, using a single actor, as in TD3, produces identical training targets for the double critics, thereby limiting the benefits they can provide \cite{pan2020softmax}.

\subsection{Representation Learning}

To enhance the learning capability of the actor and critic, DADC-R, DASC-R, and SASC-R incorporate a dedicated representation learning module consisting of a pair of encoders. The state encoder $E_s$ maps a state $s$ to a state embedding, denoted as $e_s = \text{AvgL1Norm}(E_s(s))$. The state-action encoder $E_{sa}$ jointly maps the concatenated vector $[e_s, a]$ to a state-action embedding, denoted as $e_{sa} = E_{sa}([e_s, a])$. Based on these, we define the augmented state as $[s, e_s]$ and the augmented action as $e_{sa}$.

\section{TDDR With Two Refinements}
\label{sec:Proposed Method}

\subsection{Tunable Bias Control in TDDR}

In this section, we introduce three algorithms based on novel convex combination schemes that operate on state–action pairs $(s, a)$ to enable tunable bias control. The key distinction between these algorithms and the benchmarks lies in $\psi$. We describe each formulation in detail below.

\subsubsection{DADC}
In DADC, $\psi$ is formed by a convex combination of two pessimistic estimates, both derived using the CDQ mechanism. This symmetric approach evaluates whether the optimistic exploration provided by double actors can effectively mitigate compounded underestimation bias, ensuring $\psi$ does not depend solely on a single actor. The specific computation of $\psi$ is given as:
\begin{align} \label{y_DADC}
  \psi=
  \begin{cases}
  \upsilon \min_{i=1,2}(Q_{\theta_i^{\prime}}(s^{\prime},a_1^{\prime})) \\ 
   \quad + (1-\upsilon)\min_{i=1,2}(Q_{\theta_i^{\prime}}(s^{\prime},a_2^{\prime})),~\text{if}~|\delta_1|\leq|\delta_2| \\
  \upsilon \min_{i=1,2}(Q_{\theta_i^{\prime}}(s^{\prime},a_2^{\prime})) \\
  \quad + (1-\upsilon)\min_{i=1,2}(Q_{\theta_i^{\prime}}(s^{\prime},a_1^{\prime})),~\text{if}~|\delta_1|> |\delta_2| 
  \end{cases}.
  \end{align}
Both $\min_{i=1,2}(Q_{\theta_i^{\prime}}(s^{\prime},a_1^{\prime}))$ and $\min_{i=1,2}(Q_{\theta_i^{\prime}}(s^{\prime},a_2^{\prime}))$ are susceptible to underestimation bias, yet these biases stem from distinct state-action pairs. Their convex combination facilitates mutual compensation, where the underestimation of one estimate is mitigated by the other, thereby reducing compounded underestimation.

\subsubsection{DASC}
In DASC, $\psi$ is formed by an asymmetric convex combination of a pessimistic CDQ estimate and an optimistic Q-learning estimate. Utilizing double actors, DASC investigates the trade-off between pessimistic underestimation and optimistic overestimation, leveraging Q-learning to mitigate underestimation while employing CDQ to prevent policy degradation due to overestimation. The specific computation of $\psi$ is given as:
\begin{align} \label{y_DASC}
  \psi=
    \begin{cases}
  \upsilon \min_{i=1,2}(Q_{\theta_i^{\prime}}(s^{\prime},a_1^{\prime})) \\
  \quad + (1-\upsilon)Q_{\theta_1^{\prime}}(s^{\prime},a_2^{\prime}),~\text{if}~|\delta_1|\leq|\delta_2| \\
  \upsilon \min_{i=1,2}(Q_{\theta_i^{\prime}}(s^{\prime},a_2^{\prime})) \\
  \quad + (1-\upsilon)Q_{\theta_1^{\prime}}(s^{\prime},a_1^{\prime}),~\text{if}~|\delta_1|> |\delta_2| 
    \end{cases}.
    \end{align}
Motivated by theorems 1 and 2 from \cite{lyu2022efficient}, where $\text{bias}(\psi_{Q_{\theta_1^{\prime}}(s^{\prime},a_2^{\prime})}) \geq \text{bias}(\psi_{\min_{i=1,2}(Q_{\theta_i^{\prime}}(s^{\prime},a_1^{\prime}))})$, this combination mitigates underestimation.

\subsubsection{SASC}
SASC also adopts an asymmetric combination of a CDQ and a Q-learning estimate but is constrained to actions from a single actor, serving as a critical ablation study to isolate the contribution of action diversity. The specific computation of $\psi$ is given as:
\begin{align} \label{y_SASC}
  \psi=
  \begin{cases}
  \upsilon \min_{i=1,2}(Q_{\theta_i^{\prime}}(s^{\prime},a_1^{\prime})) \\
  \quad + (1-\upsilon)Q_{\theta_1^{\prime}}(s^{\prime},a_1^{\prime}),~\text{if}~|\delta_1|\leq|\delta_2|\\
  \upsilon \min_{i=1,2}(Q_{\theta_i^{\prime}}(s^{\prime},a_2^{\prime})) \\
  \quad + (1-\upsilon)Q_{\theta_1^{\prime}}(s^{\prime},a_2^{\prime}),~\text{if}~|\delta_1|> |\delta_2| 
  \end{cases}.
  \end{align}
This design evaluates whether bias mitigation alone suffices for superior performance or if the exploratory breadth provided by double actors is essential for escaping local optima. 

\begin{remark}
The convergence of TDDR to the optimal Q-function under simultaneous and random update patterns is formally established in the work \cite{chen2025doubleactorcritic}. These three variants extend TDDR, with their core modification being the use of a linear convex combination to construct the TD target. Since a linear convex combination of convergent operators preserves convergence to the same optimal value, DADC, DASC, and SASC also converge to the optimal Q-function under standard assumptions.
\end{remark}

\subsection{Representation Learning for Performance Enhancement}

While the convex combination mechanisms detailed above effectively control estimation bias, the performance is also critically dependent on the quality of the input features. To further enhance performance, we introduce a representation learning module as a synergistic refinement.

The embeddings are split into state and state-action representations so that the encoders can be trained with a dynamics prediction loss that solely relies on the next state $s^\prime$ \cite{Fujimoto2023TD7}. As a result, the encoders are jointly trained using the mean squared error (MSE) between the state-action embedding $e_{sa}$ and the embedding of the next state $e_{s^\prime}$:
\begin{align} \label{dynamics_prediction_loss}
L(E_s, E_{sa}) = (e_{sa} - |e_{s^\prime}|_{\times})^2,
\end{align}
where $|\cdot|_{\times}$ denotes the stop-gradient operation. The encoders $(E_s, E_{sa})$ are trained online but are decoupled (gradients from the value function and policy are not propagated to $(E_s, E_{sa})$).

Similar to TD7, our framework decouples the learning of representations from the policy and value function to ensure training stability. This is achieved by maintaining fixed and target fixed encoders:
1) The encoders $(E_s^{t+1},E_{sa}^{t+1})$ are trained with \eqref{dynamics_prediction_loss}, concurrently with the RL agent, updated at the same frequency as the value function and policy.
2) The fixed encoders $(E_s^t,E_{sa}^t)$ that provides consistent inputs for $\pi_{\phi_i}$ and $Q_{\theta_i}$.
3) The target fixed encoders $(E_s^{t-1},E_{sa}^{t-1})$ that provides stable representations for $\pi_{\phi_i^{\prime}}$ and $Q_{\theta_i^{\prime}}$.
The fixed and target fixed encoders are not updated using \eqref{dynamics_prediction_loss}. Instead, every 250 steps the iteration is incremented, and the fixed and target fixed encoders are updated simultaneously:
\begin{align}\label{encoder_loss}
(E_s^{t-1}, E_{sa}^{t-1}) \leftarrow (E_s^t, E_{sa}^t),(E_s^t, E_{sa}^t) \leftarrow (E_s^{t+1}, E_{sa}^{t+1}).
\end{align}
Applying this module to our three core algorithms yields their representation-enhanced counterparts: DADC-R, DASC-R, and SASC-R. The modification is straightforward: instead of operating on $s$, the actor networks receive $[s,e_s]$ as input. Retaining $a$ as an input is critical, as $e_{sa}$ learned via \eqref{dynamics_prediction_loss} is inherently incomplete. It primarily captures information related to state transitions and may overlook other critical aspects, such as those associated with the reward function. Consequently, $a$ is preserved as an input for the critic networks. The critic networks receive $[s,e_s]$, $e_{sa}$, and $a$ as inputs. All convex combination strategies, including \eqref{y_DADC}, \eqref{y_DASC}, and \eqref{y_SASC}, remain unchanged in principle but operate on these augmented representations.

We define the TD errors for DADC-R, DASC-R, and SASC-R, adhering to the same principle as in \eqref{deltai}, but with a critical modification. The TD error for the critic targets $Q_{\theta_i^{\prime}}$ is defined with representation learning as follows:
\begin{align} \label{deltai-representation}
  \delta_{i=1,2} = & r + \gamma \min_{j=1,2}(Q_{\theta_j^{\prime}}([s^{\prime},e_{s^{\prime}}],a_i^{\prime},e_{s^{\prime}a_i^{\prime}})) \\ \nonumber
  - & \min_{j=1,2}(Q_{\theta_j^{\prime}}([s,e_s],a,e_{sa})).
\end{align}
The gradients $\nabla_a Q_{\theta_i}$, used for updating $\theta_i$ under representation learning, are obtained by minimizing:
\begin{equation}\label{Q-loss}
N^{-1}\sum_{s}(y-Q_{\theta_i}([s,e_s],a,e_{sa}))^2.
\end{equation}
The actor networks $\pi_{\phi_i}$, also incorporating representation learning, are updated by minimizing:
\begin{align}\label{Actor-loss}
  N^{-1}\sum_{s}(\nabla_a Q_{\theta_i}([s,e_s],a,e_{sa})|_{a=\pi_{\phi_i}([s,e_s])}\nabla_{\phi_i}\pi_{\phi_i}([s,e_s])).
\end{align}
The target networks $Q_{\theta_i^{\prime}}$ and $\pi_{\phi_i^{\prime}}$ are updated following conventional rules, similar to \eqref{targetupdate}, succinctly denoted as: 
\begin{align*}
\theta_i^{\prime}\leftarrow(\theta_i,\theta_i^{\prime}),\;
  \phi_i^{\prime}\leftarrow(\phi_i,\phi_i^{\prime}),
\end{align*} 
for $i=1,2$. The update of each actor networks $\pi_\phi$ is performed as $\phi_i \leftarrow  \phi_i +\alpha \nabla_{\phi_i} J(\phi_i)$
where $\nabla_{\phi_i} J(\phi_i)$ is defined in \eqref{Actor-loss} using the corresponding critic network $Q_{\theta_i}$. 

\begin{figure*}[t]
  \centering
  \includegraphics[width=0.76\textwidth]{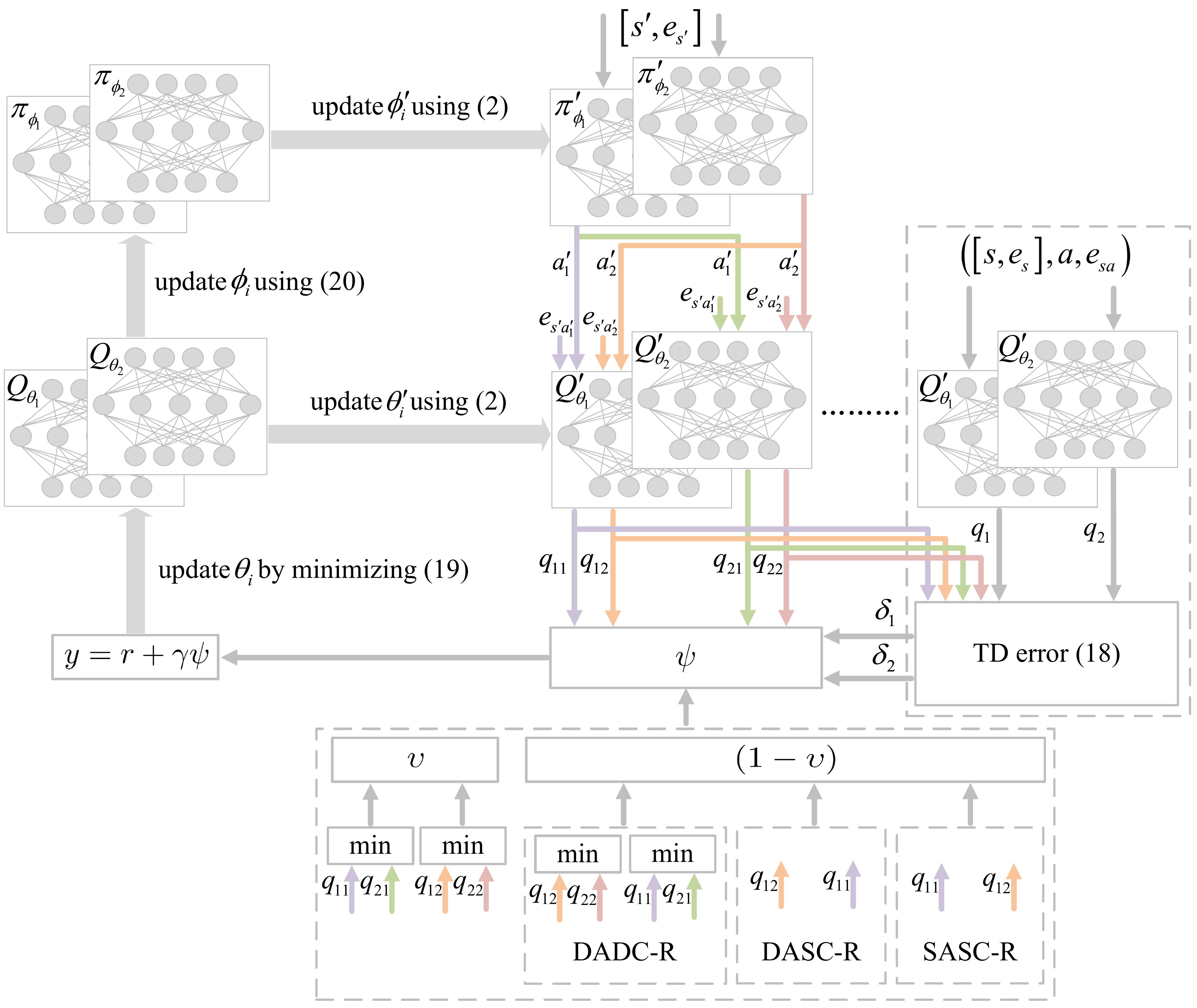}
  \caption{Architecture of DADC-R, DASC-R, and SASC-R: $q_{ij} = Q_{\theta_i^\prime}([s^\prime,e_{s^\prime}],a_j^\prime,e_{s^\prime a_j^\prime})$ and $q_i = Q_{\theta_i^\prime}([s,e_s],a,e_{sa})$; the duplication of $Q_{\theta_1^\prime}/Q_{\theta_2^\prime}$ indicates that the same networks are used with different inputs; the action $a_i'$ is generated by $\pi_{\phi_i^\prime}([s^{\prime},e_{s^\prime}])$; the notation $[\cdot]$ represents the concatenation of vectors along the feature dimension; ``min'' means a minimum of two values. The input $\psi$ is partitioned into two components, with $\upsilon$ representing the common part across all algorithms, and $1-\upsilon$ corresponding to the algorithm-specific part. When $|\delta_1| \leq |\delta_2|$, DADC-R combines $\upsilon\min(q_{11}, q_{21})$ and $(1-\upsilon) \min(q_{12}, q_{22})$. Conversely, when $|\delta_1| > |\delta_2|$, DADC-R combines $\upsilon \min(q_{12}, q_{22})$ and $(1-\upsilon) \min(q_{11}, q_{21})$. DASC-R and SASC-R adhere to the same computational rules.}
  \label{fig:TDDRvariants}
  \end{figure*}

In summary, we present DADC-R, DASC-R, and SASC-R, as detailed in Algorithm~\ref{alg:TDDR_via_Convex_Combination}, with their architectures illustrated in Fig.~\ref{fig:TDDRvariants}.

\subsection{Comparison with TD7 in Representation Learning}

Although DADC-R, DASC-R, SASC-R, and TD7 employ the same encoder architecture to generate $e_s$ and $e_{sa}$, their fundamental distinction lies in the utilization of these embeddings. DADC-R, DASC-R, and SASC-R use a three-layer network, whereas TD7 employs a four-layer network.

For critic networks, DADC-R, DASC-R, and SASC-R adopt a unified input strategy, concatenating $\big[ [s, e_s], a, e_{sa} \big]$ at the first layer for forward propagation. In TD7, the critic networks first normalize the state-action pair using $\text{AvgL1Norm}[s, a]$ at the first layer, then concatenate the learned embeddings $[e_s, e_{sa}]$, and finally concatenate these into $[\text{AvgL1Norm}[s, a], e_s, e_{sa}]$ at the second layer for forward propagation.

For actor networks, DADC-R, DASC-R, and SASC-R use the augmented state $[s, e_s]$ as the only input. In TD7, the actor network first normalizes the state with $\text{AvgL1Norm}(s)$ at the first layer, then concatenates $[\text{AvgL1Norm}(s), e_s]$ at the second layer for forward propagation.

Increasing the dimensionality of the state-action input to the value function will cause extrapolation error \cite{Fujimoto2023TD7}. In contrast, TD7 mitigates this issue by explicitly clipping the learned Q-values within the range of the maximum and minimum values observed in the dataset. Meanwhile, DADC-R, DASC-R, and SASC-R leverage \eqref{deltai-representation} to dynamically select the actor target network that yields a smaller TD error for computing next actions in the TD target, thereby achieving stability without imposing explicit constraints on the TD target. Notably, our representation learning approach is developed within the DAC framework, whereas TD7 is formulated under the AC framework.

\begin{algorithm}[tb]
    \caption{DADC-R, DASC-R, SASC-R}
    \label{alg:TDDR_via_Convex_Combination}
    \begin{algorithmic}[1] 
    \STATE Initialize critic networks $Q_{\theta_1},Q_{\theta_2}$ and actor networks $\pi_{\phi_1},\pi_{\phi_2}$ with random parameters $\theta_1,\theta_2,\phi_1,\phi_2$ 
    \STATE Initialize target networks $\theta_1^{\prime},\theta_2^{\prime},\phi_1^{\prime},\phi_2^{\prime}$ and experience replay buffer $\mathfrak{R}$
    \STATE Initialize encoders $E_s^{t+1},E_{sa}^{t+1}$, fixed encoders $E_s^t,E_{sa}^t$, and target fixed encoders $E_s^{t-1},E_{sa}^{t-1}$
    \FOR{$t$ = 1 to $T$}
    \STATE Calculate $e_s = \text{AvgL1Norm}(E_s^t(s))$, $a_j = \pi_{\phi_j}([s,e_s])$, $e_{sa_j} = E_{sa}^t([e_s, a_j])$
    \STATE Select action $a$ with $\max_{i,j} Q_{\theta_i}([s,e_s],a_j,e_{sa_j})$
    \STATE Execute action $a$ and observe reward $r$, new state $s^{\prime}$ and done flag $d$
    \STATE Store transitions in the experience replay buffer, i.e., $(s_t,a_t,r_t,s_{t+1}^{\prime},d_t)$
    \FOR{$i = 1,2$}
    \STATE Sample $\{(s_t,a_t,r_t,s_{t+1}^{\prime},d_t)\}_{t=1}^{T}\sim\mathfrak{R}$
    \STATE Calculate $e_{s^{\prime}} = \text{AvgL1Norm}(E_s^{t-1}(s^{\prime}))$, $a_i^{\prime} = \pi_{\phi_i^{\prime}}([s^{\prime},e_{s^{\prime}}])+\epsilon,\epsilon\sim$ clip${(\mathcal{N}(0,\bar{\sigma}),-c,c)}$, $e_{s^{\prime}a_i^{\prime}} = E_{sa}^{t-1}([e_{s^{\prime}}, a_i^{\prime}])$
    \STATE Train ($E_s^{t+1},E_{sa}^{t+1}$) with \eqref{dynamics_prediction_loss}
    \STATE Calculate $\delta_i$ with \eqref{deltai-representation}
    \STATE Calculate $y$ with \eqref{TDpsi}, \eqref{y_DADC}, \eqref{y_DASC}, and \eqref{y_SASC}, incorporating representation learning
    \STATE Update critic $\theta_i$ by minimizing \eqref{Q-loss}
    \STATE Update actor $\phi_i$ with policy gradient \eqref{Actor-loss}
    \STATE Update target networks: \\ $\theta_i^{\prime}\leftarrow(\theta_i,\theta_i^{\prime}),\ \phi_i^{\prime}\leftarrow(\phi_i,\phi_i^{\prime})$
    \IF{$t \bmod 250 == 0$}
    \STATE Update ($E_s^t,E_{sa}^t$) and ($E_s^{t-1},E_{sa}^{t-1}$) with \eqref{encoder_loss}
    \ENDIF
    \ENDFOR
    \ENDFOR
    \end{algorithmic}
\end{algorithm}

\section{Experiments}
\label{sec:Experiments}

We adopt two widely-used continuous benchmarks,  OpenAI Gym \cite{brockman2016openaigym} simulated by MuJoCo \cite{todorov2012mujoco} and Box2d \cite{catto2011box2d} to test the performance of our methods against benchmarks. Our benchmarks include DDPG \cite{lillicrap2019continuouscontroldeepreinforcement}, TD3 \cite{fujimoto2018addressing}, DARC \cite{lyu2022efficient}, SD3 \cite{pan2020softmax}, GD3 \cite{lyu2023value}, TDDR \cite{chen2025doubleactorcritic}, SAC \cite{haarnoja2018softSAC}, and PPO \cite{schulman2017proximalPPO}. Each of the selected environments was run for 1 million steps, with evaluations conducted every 5,000 steps using the average reward from 10 episodes for each evaluation.

This section is organized into five parts. Section~\ref{subsec:Ablation Study} conducts an ablation study to evaluate the contributions of individual components in DADC-R, DASC-R, and SASC-R, identifying their impact on overall performance. Section~\ref{subsec:Mitigating Underestimation Bias} analyzes the efficacy of DADC, DASC, and SASC in mitigating estimation bias. Section~\ref{subsec:Extensive Experiments} presents a comprehensive comparison of DADC-R, DASC-R, and SASC-R against benchmark algorithms, demonstrating their competitive advantage. Finally, Section~\ref{subsec:SOTA} compares DADC-R, DASC-R, and SASC-R with SOTA algorithms, further validating their superior performance. 

The values of the hyperparameters are listed in Tables~\ref{tab:all_hyperparameters} and \ref{tab:upsilon}. For BipedalWalker and InvertedDoublePendulum, we set $\upsilon=1$ for all TDDR variants. It should be noted that, for a fair comparison, the hyperparameters of DARC, SD3, and GD3 were selected based on the results of ablation experiments from their original papers.

\begin{table}[h!] 
  \caption{Hyperparameters setup
  \label{tab:all_hyperparameters}}
  \centering
  \small
  \begin{tabular}{cc}
    \toprule
    Hyperparameter & Value \\
    \midrule
    \textbf{DARC} \\
    regularization parameter $\lambda$ & 0.005 \\
    weighting coefficient $\nu$ & 0.1 \\
    \textbf{SD3} \\
    noise samples & 50 \\
    parameter $\beta$ & 0.001 \\
    \textbf{GD3} \\
    noise samples & 50 \\
    parameter $\beta$ & 0.05 \\
    bias term $b$ & 2 \\
    activation functions & 
    environment dependent\\
   \bottomrule
  \end{tabular}
\end{table}


\begin{table}[!t] 
    \caption{Hyperparameter $\upsilon$ setup for each algorithm
    \label{tab:upsilon}}
    \centering
    \begin{tabular}{ccccc}
       \toprule
       Environment & Ant & HalfCheetah & Hopper & Walker2d \\
       \midrule
       DADC   & 0.3 & 0.0 & 0.9 & 0.4 \\
       DASC   & 0.5 & 0.0 & 0.9 & 0.8 \\
       SASC   & 0.0 & 0.2 & 0.8 & 0.0 \\
       DADC-R & 0.6 & 0.3 & 0.6 & 0.2 \\
       DASC-R & 0.9 & 0.1 & 0.4 & 0.4 \\
       SASC-R & 0.8 & 0.8 & 0.8 & 0.8 \\
       \bottomrule
    \end{tabular}
\end{table}

\subsection{Ablation Study}
\label{subsec:Ablation Study}

\subsubsection{Component Analysis}

DADC-R, DASC-R, and SASC-R are built upon four key components: 1) a representation learning module, 2) the DA-CDQ mechanism, 3) the TDDR framework, and 4) a convex combination strategy. Removing either actor disrupts the DA-CDQ mechanism and invalidates \eqref{deltai-representation}.

To analyze the contribution of each component, we design the ablation studies as follows:

a) We analyze DADC, DASC, SASC, DADC-R, DASC-R, and SASC-R, enabling direct comparison and isolating the impact of the representation learning component.

b) Removing the convex combination strategy (or setting $\upsilon = 1$), DADC, DASC, and SASC alongside DADC-R, DASC-R, and SASC-R, reduce to TDDR without and with representation learning, respectively.

c) Removing one actor reduces TDDR to TD3 \cite{chen2025doubleactorcritic}.

\subsubsection{Performance Comparison}

\begin{figure*}
  \centering
  \subcaptionbox{}{\includegraphics[width = 0.24\textwidth]{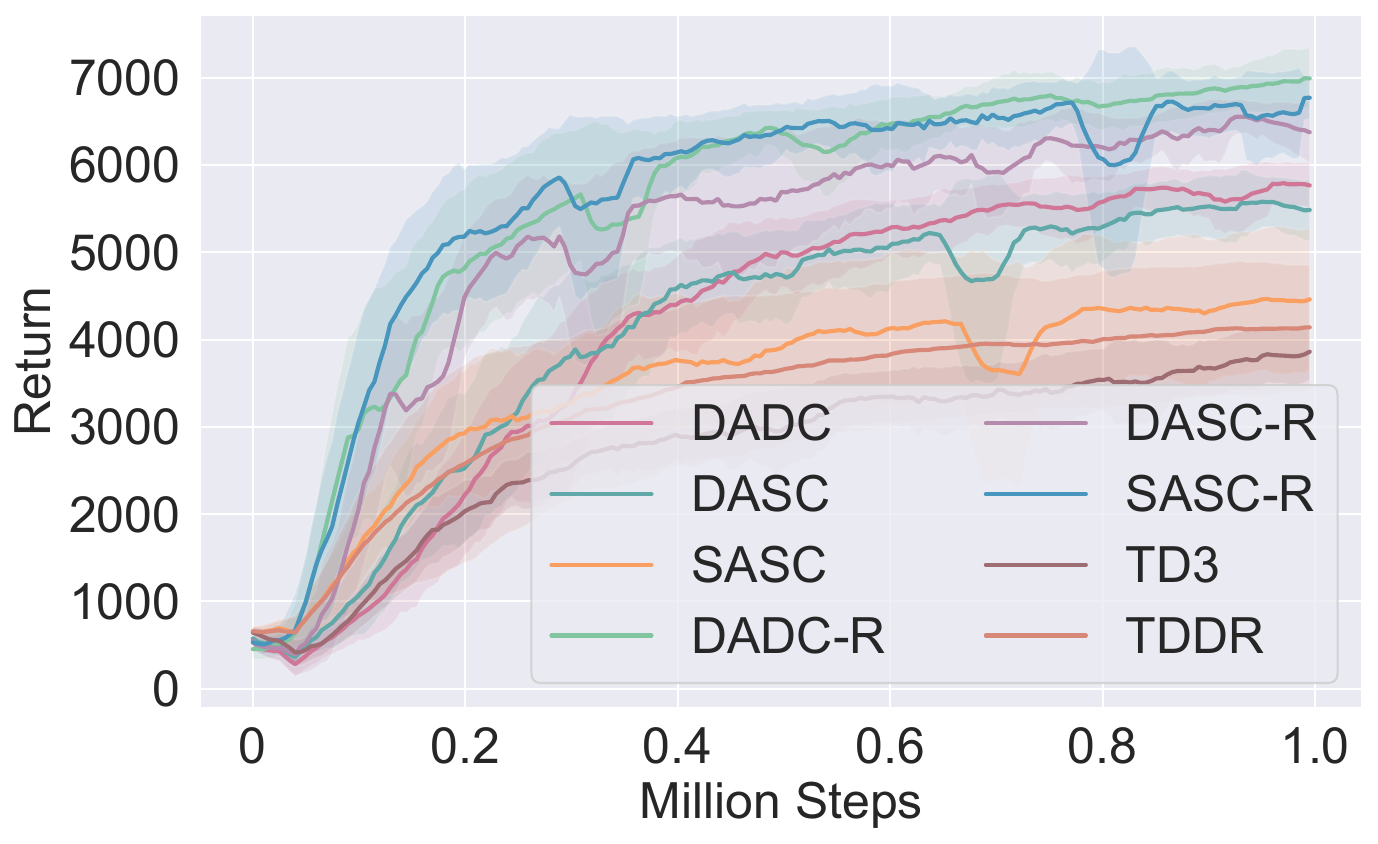}}
  \hfill
  \subcaptionbox{}{\includegraphics[width = 0.24\textwidth]{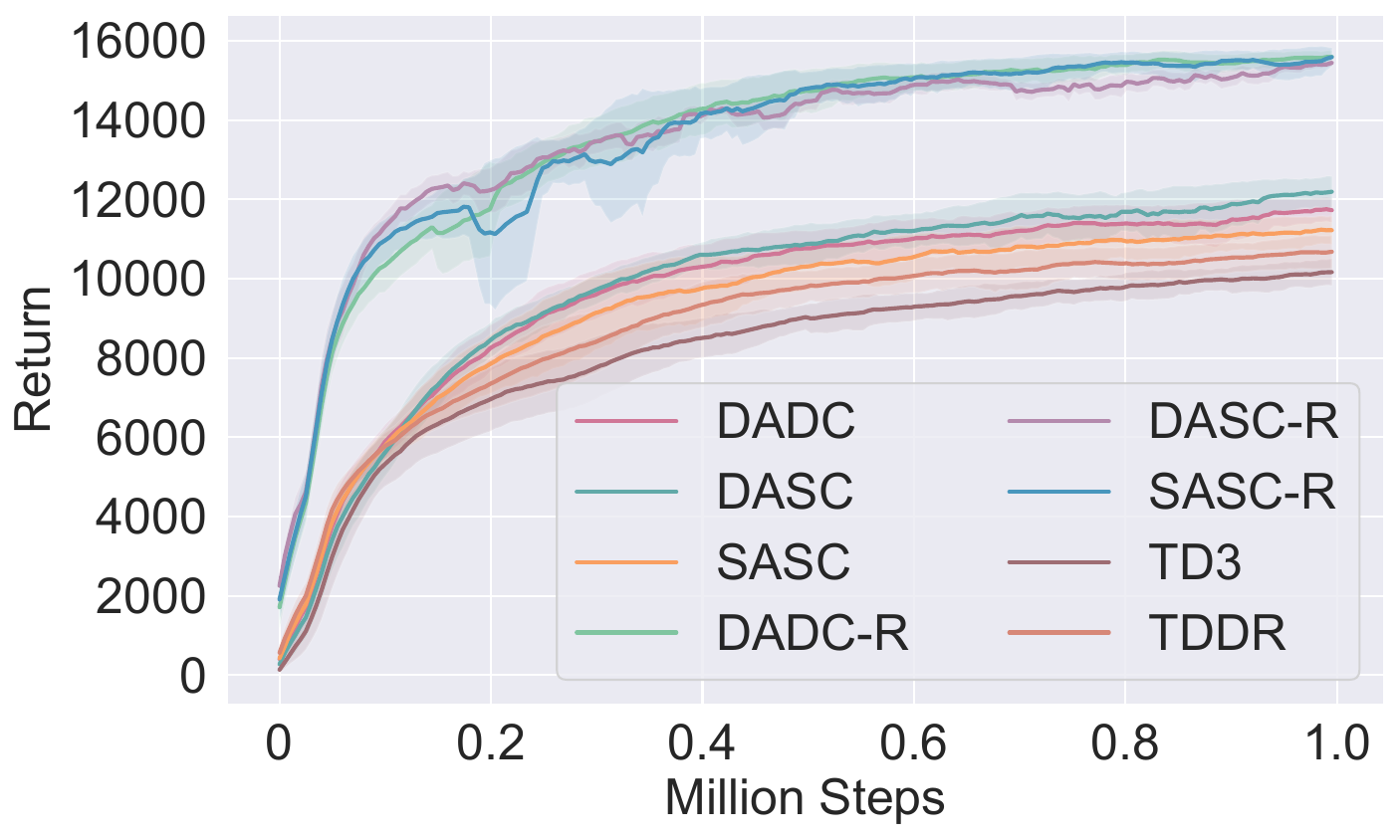}}
  \hfill
  \subcaptionbox{}{\includegraphics[width = 0.24\textwidth]{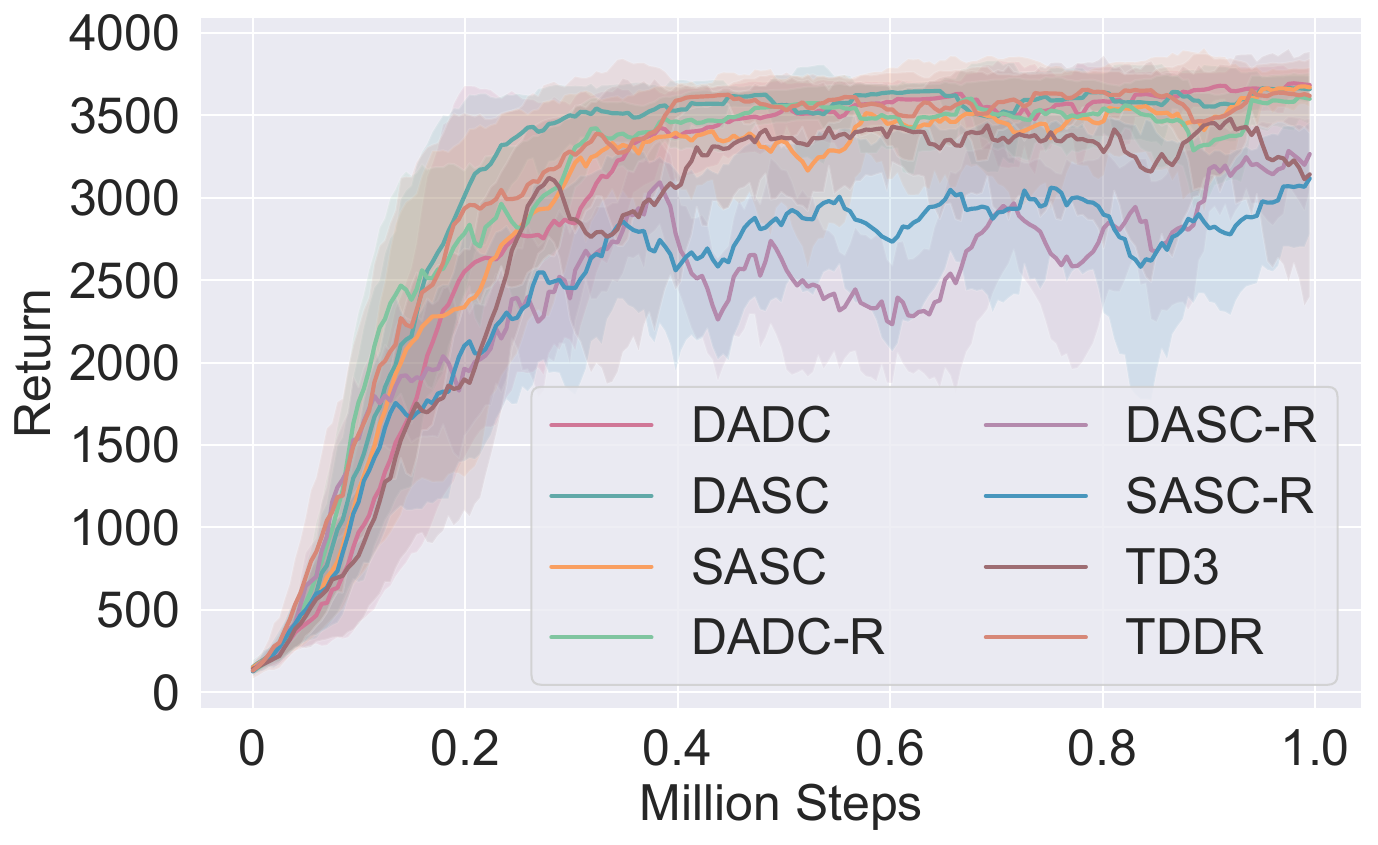}}
  \hfill
  \subcaptionbox{}{\includegraphics[width = 0.24\textwidth]{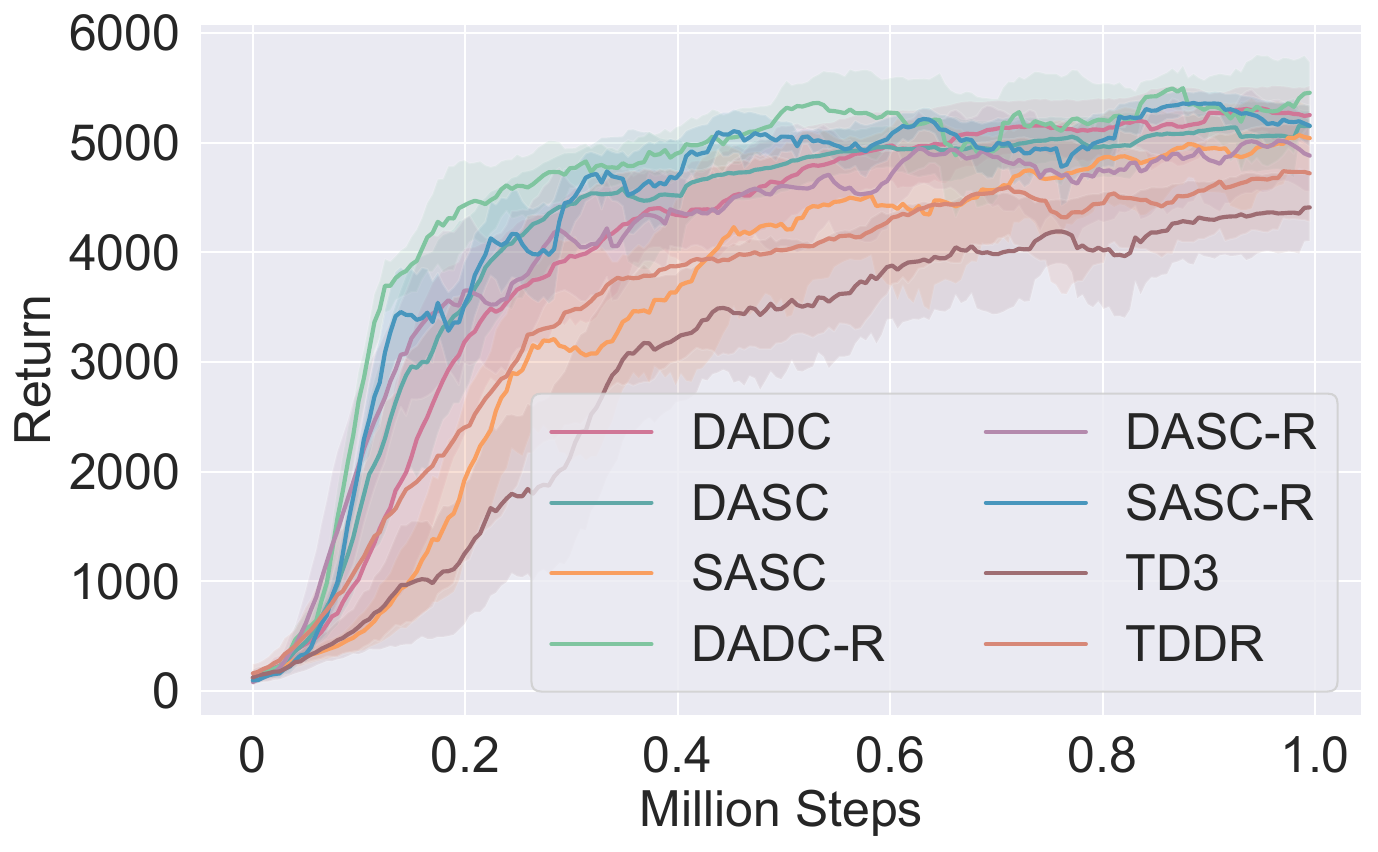}}
  \hfill
  \caption{Comparison of DADC, DASC, SASC, DADC-R, DASC-R, SASC-R, TD3 and TDDR across four environments. (a) Ant-v2, (b) HalfCheetah-v2, (c) Hopper-v2, (d) Walker2d-v2.}
  \label{fig:TD3_TDDR_threeTDDRvariants}
\end{figure*}

The overall performance is plotted in Fig. \ref{fig:TD3_TDDR_threeTDDRvariants}, where the solid curves depict the mean across evaluations and the shaded region represents one standard deviation over five runs. A sliding window of five was applied to achieve smoother curves. This same representation is also used in Fig. \ref{fig:mitigate_bias_compareTD3}.

Table~\ref{tab:numerical_comparison_baseline_algorithm} displays the average returns from five random seeds for the last ten evaluations, with the maximum value for each task highlighted in bold. This formatting is also applied in Tables \ref{tab:numerical_comparison_algorithm} and \ref{table.SOTA}.

As illustrated in Fig.~\ref{fig:TD3_TDDR_threeTDDRvariants} and Table~\ref{tab:numerical_comparison_baseline_algorithm}, DADC-R, DASC-R, and SASC-R consistently outperform TD3 and TDDR across three environments. Similarly, DADC, DASC, and SASC demonstrate superior performance compared to TD3 and TDDR across four environments. However, DADC, DASC, and SASC exhibit significant performance gaps compared to DADC-R, DASC-R, and SASC-R in Ant and HalfCheetah. Performance also declines in Walker2d, except for DASC, underscoring the critical role of the representation learning module in enhancing performance.

\begin{table*}
  \caption{Average return and standard deviation for Fig. \ref{fig:TD3_TDDR_threeTDDRvariants}
  \label{tab:numerical_comparison_baseline_algorithm}}
  \centering
  \begin{tabular}{ccccccccccccccc}
    \toprule
    Algorithms & DADC & DASC & SASC & DADC-R & DASC-R & SASC-R & TD3 & TDDR \\
    \midrule
      Ant & 5779$\pm$122 & 5524$\pm$133 & 4444$\pm$817 & $\bm{6958 \pm 118}$ & 6452  $\pm$ 347 & 6590  $\pm$ 618 & 3811$\pm$213 & 4129$\pm$726 \\
      HalfCheetah & 11694$\pm$208 & 12137$\pm$240 & 11187$\pm$284 & 15570  $\pm$ 189 & $\bm{15672 \pm 152}$   & 15467  $\pm$ 292 & 10105$\pm$208 & 10666$\pm$392 \\
      Hopper & $\bm{3691\pm75}$ & 3665$\pm$46 & 3660$\pm$109 & 3581  $\pm$ 124 & 3284  $\pm$ 411 & 3070  $\pm$ 474 & 3197$\pm$282 & 3631$\pm$160 \\
      Walker2d & 5272$\pm$193 & 5050$\pm$244 & 5046$\pm$266 & $\bm{5359\pm427}$ & 5000  $\pm$ 408 &  5189  $\pm$ 291 & 4360$\pm$229 & 4733$\pm$423 \\
    \bottomrule
  \end{tabular}
\end{table*}

\subsection{Analysis of Bias Mitigation}
\label{subsec:Mitigating Underestimation Bias}

In this section, we first evaluate the estimation biases of DADC, DASC, and SASC across various settings of $\upsilon$. Subsequently, we compare the estimation biases of DADC, DASC, and SASC against those of TD3 and TDDR.

\begin{figure}[t]
  \centering
  \includegraphics[width=0.48\textwidth]{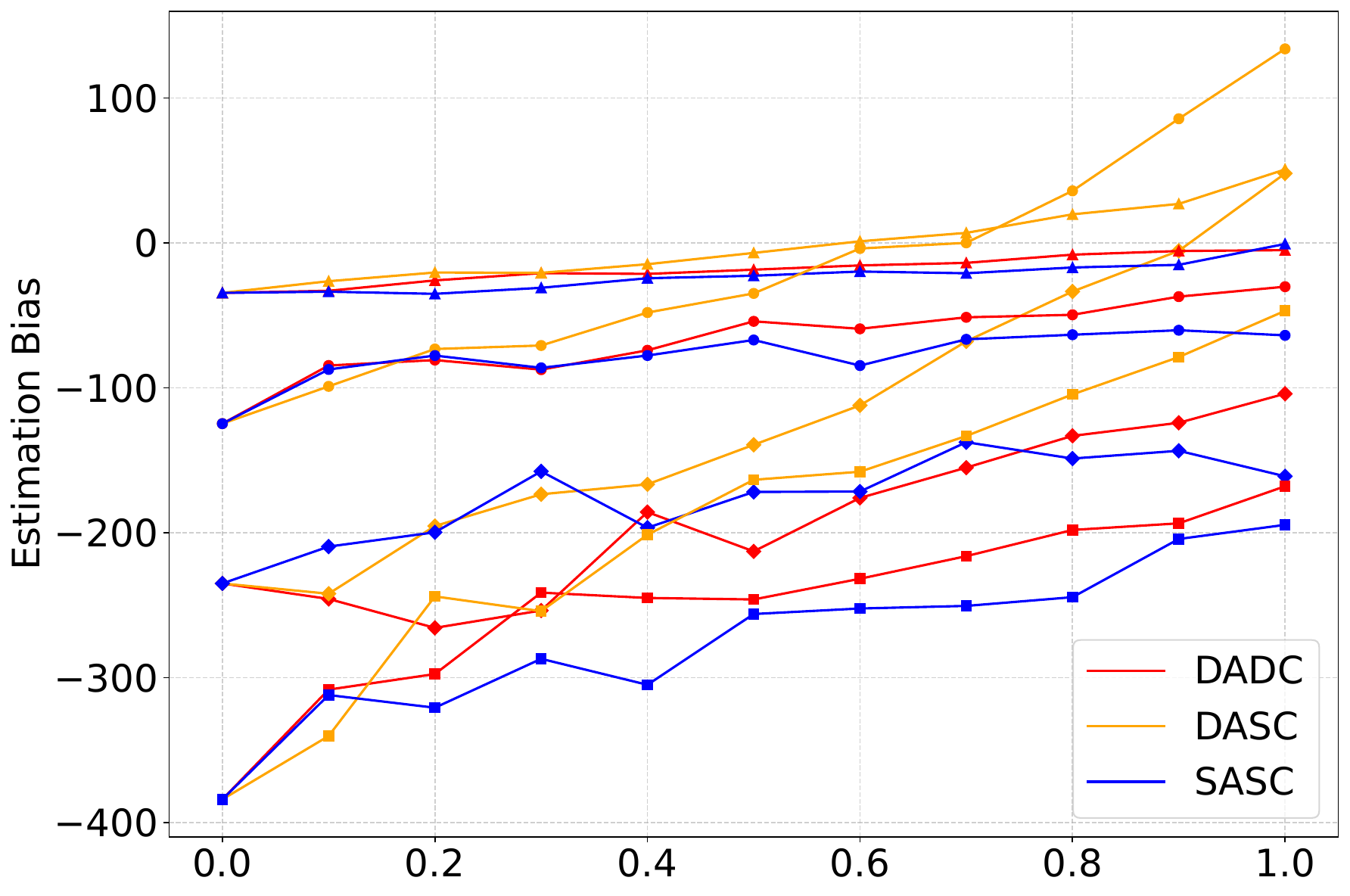}
  \caption{Comparison of estimation bias for DADC, DASC, and SASC across four environments. Markers denote distinct environments: diamonds for Ant, squares for HalfCheetah, triangles for Hopper, and circles for Walker2d.}
  \label{fig:threeTDDRvariants}
  \end{figure}

Fig.~\ref{fig:threeTDDRvariants} depicts the estimation bias of each algorithm under varying hyperparameter settings in Ant, HalfCheetah, Hopper, and Walker2d. Each point represents the average estimation bias over five random seeds for the last ten evaluations. The horizontal axis corresponds to the values of $\upsilon$. As shown in Fig.~\ref{fig:threeTDDRvariants}, estimation bias consistently increases as $\upsilon$ decreases, transitioning from underestimation to overestimation.

The following key observations can be made from Fig. \ref{fig:threeTDDRvariants}.

First, the architectural design of each variant governs its sensitivity to bias modulation, revealing a clear hierarchy in their progression from pessimism to optimism as $\upsilon$ decreases. Experimental results consistently show that this transition occurs fastest in DASC, followed by DADC, and slowest in SASC. The rapid shift in DASC is attributable to its asymmetric integration of a Q-learning estimate and the inherent exploratory capacity of double actors. The most insightful comparison, however, lies between DADC and SASC. Although SASC explicitly includes an optimistic Q-learning estimate, its transition toward overestimation is slower than that of DADC, which relies on double actors. This observation highlights that exploration is a more potent mechanism for inducing optimism than a standard Q-learning component.

Second, superior exploration is critical in complex environments like Ant. The Ant environment, with its high-dimensional state-action space (111, 8), poses a significant exploration challenge. Here, the architectural differences become starkly apparent. As shown in Table \ref{tab:numerical_comparison_baseline_algorithm}, SASC exhibits a significant performance gap compared to DADC and DASC, even with an optimally tuned $\upsilon$. This demonstrates that in exploration-heavy tasks, effective bias control alone is insufficient without a powerful exploration mechanism.

Third, the optimal bias strategy is environment-dependent, and some environments are more tolerant of overestimation. In HalfCheetah, the optimal performance for DADC and DASC is achieved at $\upsilon=0$, corresponding to overestimation relative to TD3, as shown in Tables \ref{tab:numerical_comparison_baseline_algorithm} and \ref{tab:td3ddpg_bias_Appendix}. This overestimation bias, however, leads to superior returns, reinforcing that not all environments require overestimation or underestimation to achieve optimal performance \cite{kuznetsov2022automatingcontroloverestimationbias}. This is also mentioned in maxmin Q-learning \cite{lan2021maxminqlearningcontrollingestimation}, where overestimation or underestimation is related to exploration, facilitating optimistic and pessimistic exploration, respectively. Both overestimation and underestimation have their advantages and disadvantages for different environments.

\begin{figure*}
  \centering
  \subcaptionbox{}{\includegraphics[width = 0.24\textwidth]{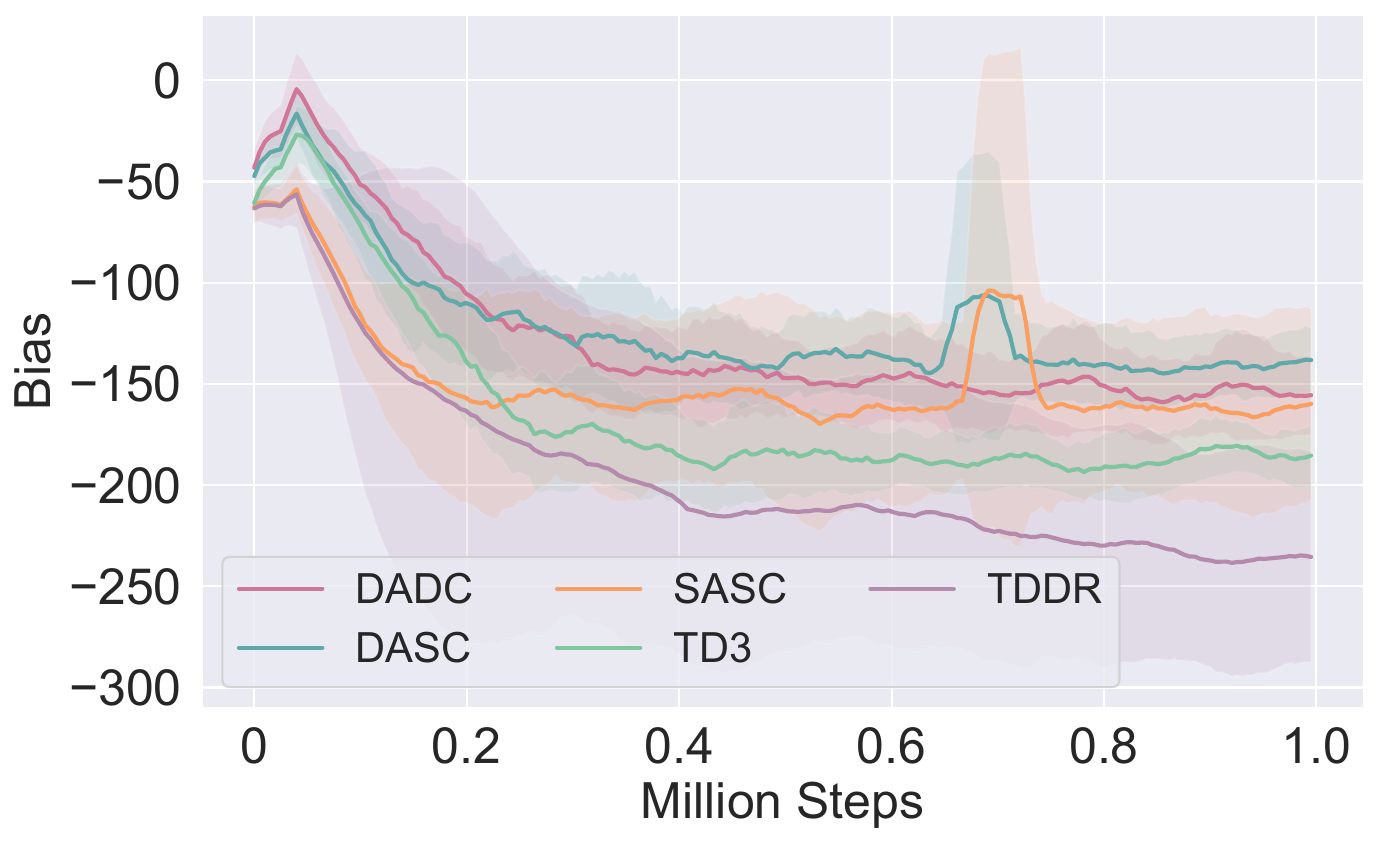}}
  \hfill
  \subcaptionbox{}{\includegraphics[width = 0.24\textwidth]{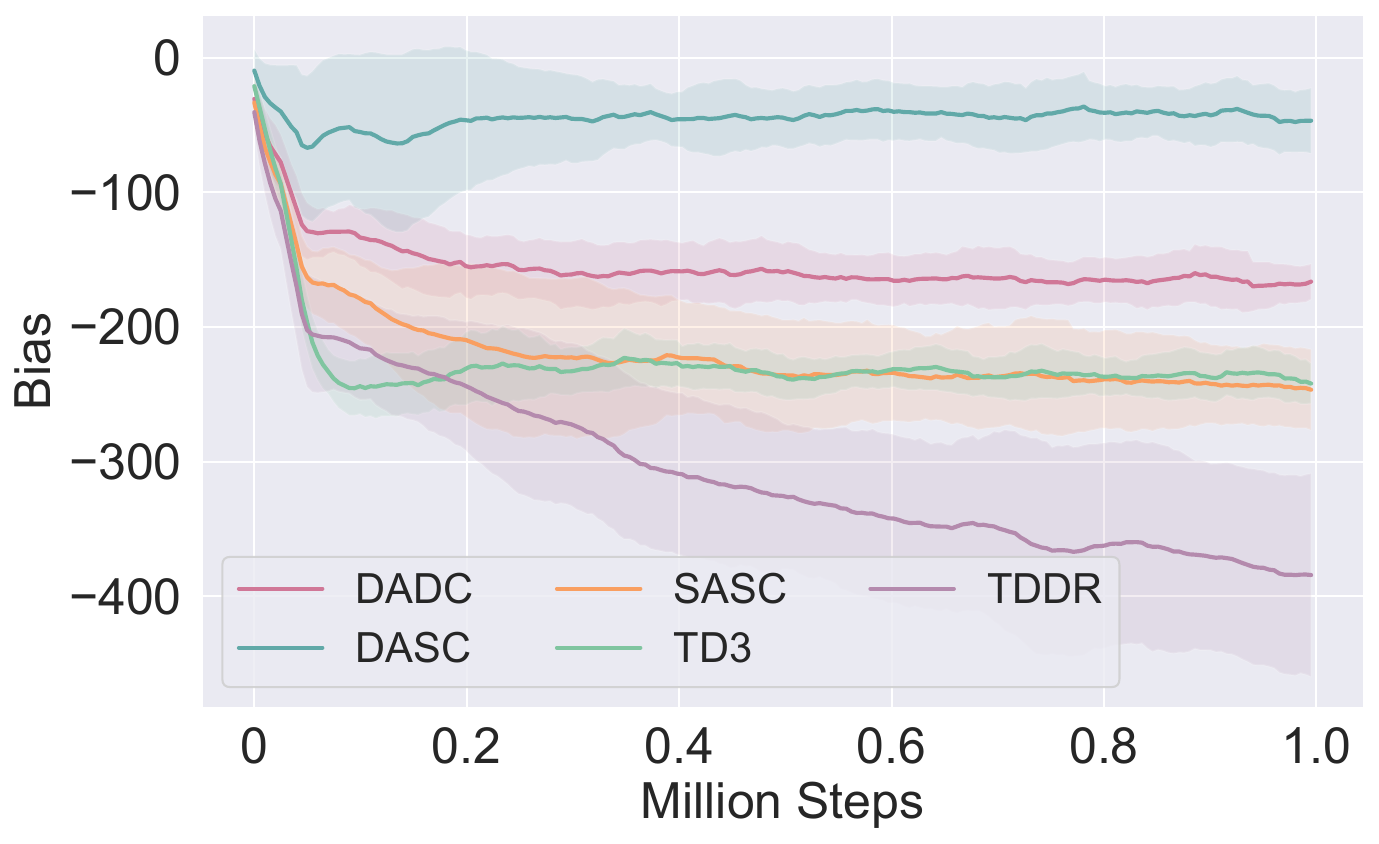}}
  \hfill
  \subcaptionbox{}{\includegraphics[width = 0.24\textwidth]{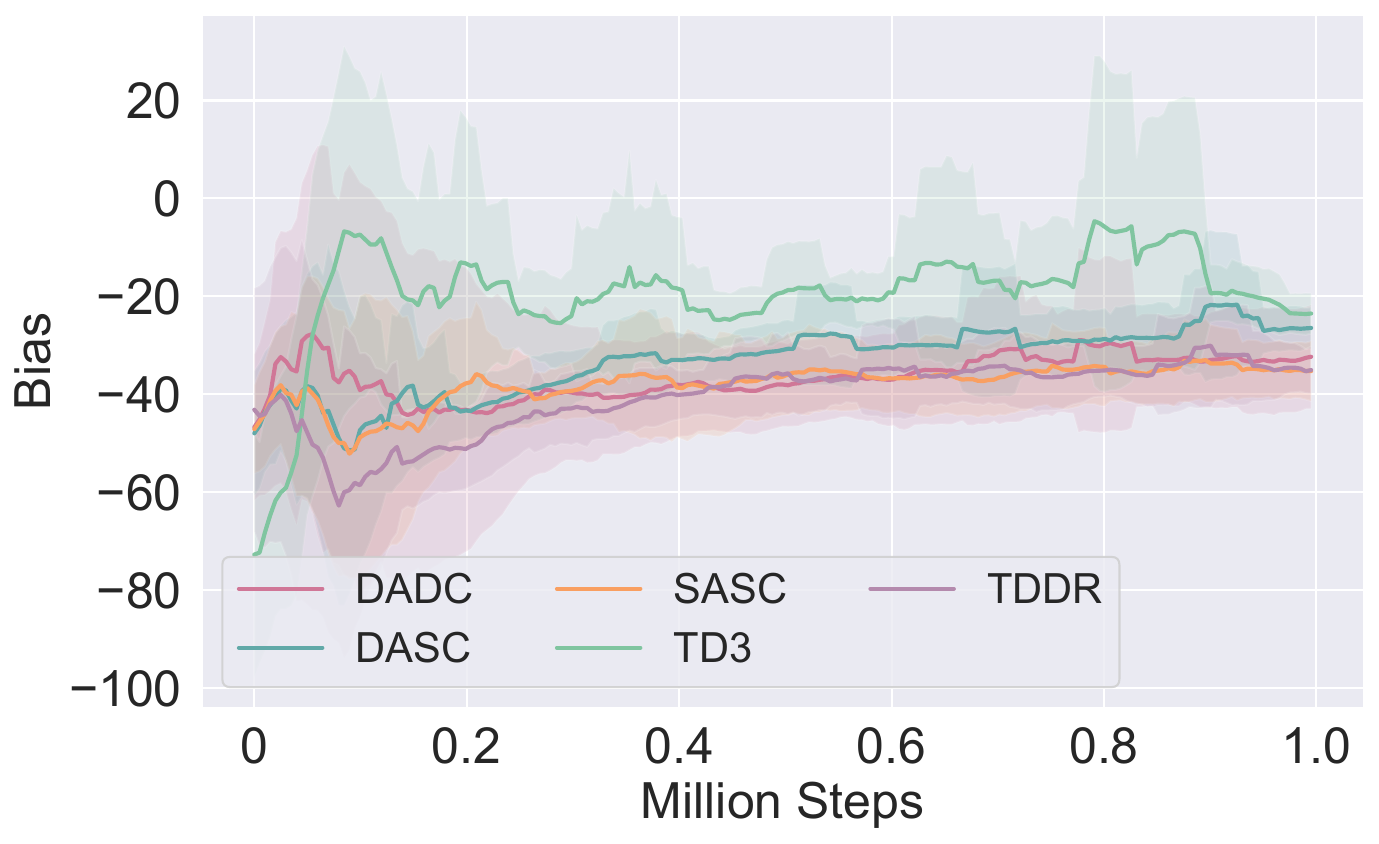}}
  \hfill
  \subcaptionbox{}{\includegraphics[width = 0.24\textwidth]{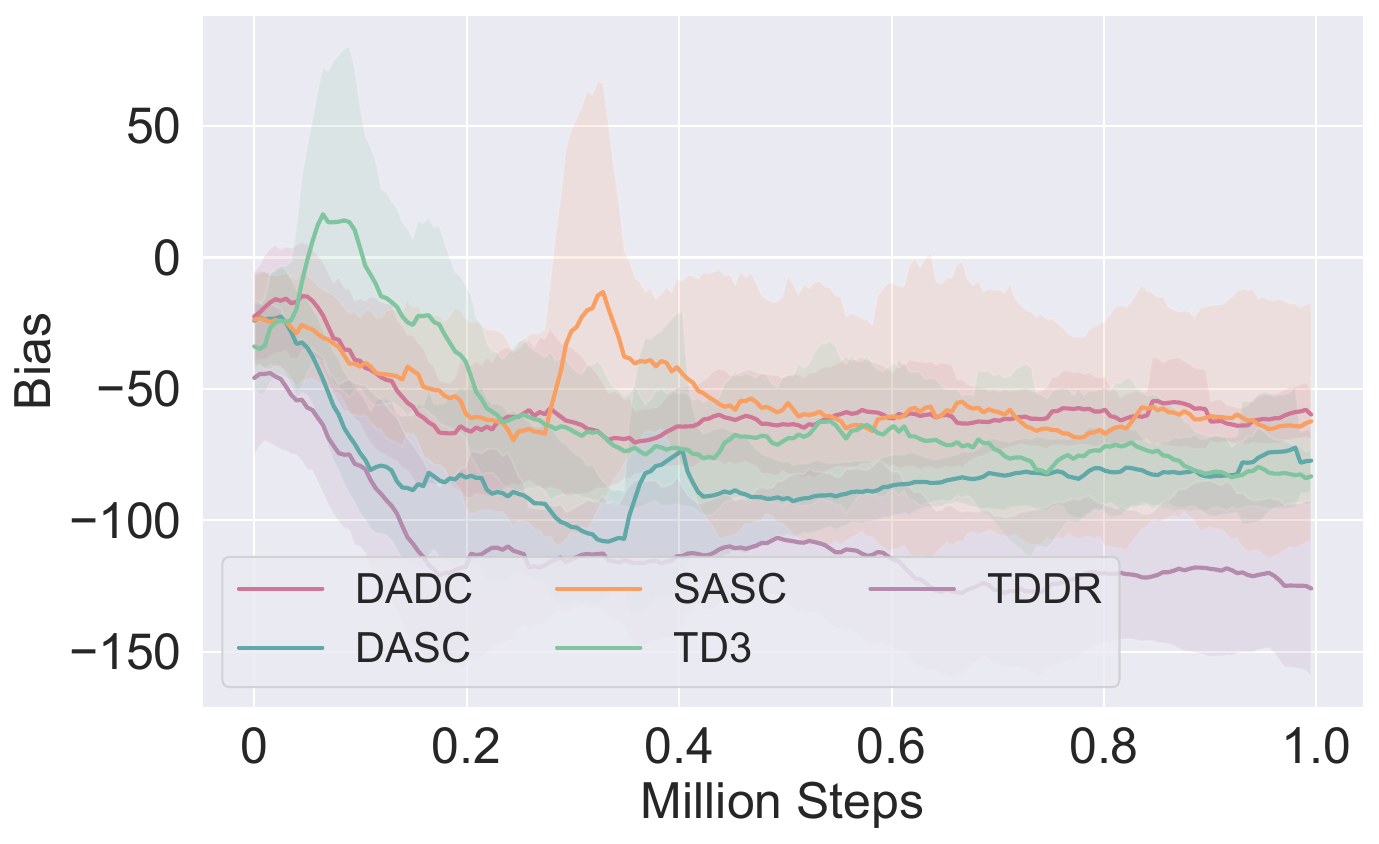}}
  \hfill
  \caption{Comparison of estimation bias among DADC, DASC, SASC, TD3, and TDDR across four environments. (a) Ant-v2, (b) HalfCheetah-v2, (c) Hopper-v2, (d) Walker2d-v2.}
  \label{fig:mitigate_bias_compareTD3}
\end{figure*}

The experimental results are shown in Fig.~\ref{fig:mitigate_bias_compareTD3}. Table~\ref{tab:td3ddpg_bias_Appendix} summarizes the results from Fig.~\ref{fig:mitigate_bias_compareTD3}, reporting the average estimation bias over five random seeds for the last ten evaluations.

As demonstrated in Fig. \ref{fig:mitigate_bias_compareTD3} and Table \ref{tab:td3ddpg_bias_Appendix}, DADC, DASC, and SASC effectively mitigate underestimation bias in Ant compared to TD3 and TDDR. Similarly, DADC and DASC reduce underestimation bias in HalfCheetah and Walker2d relative to TD3 and TDDR. Although DADC exhibits some residual underestimation bias in Hopper, it consistently achieves higher returns than TD3 and TDDR.

\begin{table}[!t]
  \caption{Estimation bias for Fig. \ref{fig:mitigate_bias_compareTD3}}  \label{tab:td3ddpg_bias_Appendix}
  \centering
  \begin{tabular}{cccccccccc}
    \toprule
    Algorithms & DADC & DASC & SASC & DDPG & TD3 & TDDR \\
    \midrule
      Ant & -155 & -139 & -161 & 428 & -187 & -235 \\
      HalfCheetah & -168 & -47 & -245 & 102 & -238 & -384 \\
      Hopper & -33 & -27 & -35 & 315 & -23 & -35 \\
      Walker2d & -59 & -73 & -64 & 391 & -82 & -125 \\
    \bottomrule
  \end{tabular}
\end{table}


\subsection{Extensive Experiments}
\label{subsec:Extensive Experiments}

The comparison is straightforward. Table \ref{tab:numerical_comparison_algorithm} demonstrates that DADC-R, DASC-R, and SASC-R consistently achieve higher average returns than benchmarks across all evaluated environments except Hopper and InvertedDoublePendulum. Notably, for BipedalWalker, we incorporate the state $s$ as an additional input to both the actor and critic networks.

For algorithms within the AC framework, DADC-R surpasses DDPG, SAC, and PPO in all environments, except for PPO in InvertedDoublePendulum. Meanwhile, DASC-R and SASC-R surpass DDPG, SAC, and PPO in all environments but fall short of SAC in Hopper and PPO in InvertedDoublePendulum.

For algorithms within the DAC framework, DADC-R outperforms DARC, SD3, and GD3 across all environments. Although DASC-R and SASC-R exhibit lower performance in Hopper compared to other DAC-based algorithms, they demonstrate superior performance in other environments.

\begin{table*}
  \caption{Average return and standard deviation for DADC-R, DASC-R, SASC-R and benchmark algorithms
  \label{tab:numerical_comparison_algorithm}}
  \centering
  \resizebox{\textwidth}{!}{
  \begin{tabular}{ccccccccccccccc}
    \toprule
    Algorithms & DADC-R & DASC-R & SASC-R & DDPG & DARC & SD3 & GD3 & SAC & PPO \\
    \midrule
      Ant & $\bm{6958 \pm 118}$ & 6452  $\pm$ 347 & 6590  $\pm$ 618 & 641$\pm$207 & 5438$\pm$215 & 4734$\pm$605 & 4990$\pm$242 & 4102$\pm$760 & 2813$\pm$130 \\
      HalfCheetah & 15570  $\pm$ 189 & $\bm{15672 \pm 152}$   & 15467  $\pm$ 292 & 9958$\pm$569 & 11297$\pm$204 & 10637$\pm$280 & 11032$\pm$523 & 10128$\pm$234 & 1649$\pm$12 \\
      Hopper & $\bm{3581 \pm 124}$ & 3284  $\pm$ 411 & 3070  $\pm$ 474 & 1712$\pm$168 & 3510$\pm$114 & 3386$\pm$162 & 3474$\pm$114 & 3356$\pm$343 & 2660$\pm$367 \\
      Walker2d & $\bm{5359\pm427}$ & 5000  $\pm$ 408 &  5189  $\pm$ 291 & 2063$\pm$231 & 4901$\pm$244 & 4672$\pm$241 & 4856$\pm$350 & 4514$\pm$345 & 2646$\pm$186 \\
      BipedalWalker & $\bm{312 \pm 9}$ & $\bm{312 \pm 9}$ & $\bm{312 \pm 9}$ & 113$\pm$38 & 307$\pm$3 & 306$\pm$5 & 305$\pm$2 & 292$\pm$39 & 263$\pm$20 \\
      InvertedDoublePendulum & 9345 $\pm$ 17 & 9345 $\pm$ 17 & 9345 $\pm$ 17 & 8279$\pm$552 & 9302$\pm$31 & 9317$\pm$34 & 9314$\pm$33 & 9294$\pm$60 & $\bm{9354\pm2}$ \\
    \bottomrule
  \end{tabular}
  }
\end{table*}

\subsection{Comparison with SOTA Algorithms}
\label{subsec:SOTA}


We further evaluate DADC-R, DASC-R, and SASC-R against several SOTA algorithms to demonstrate its overall performance. We select TD3+SALE \cite{Fujimoto2023TD7}, TD7 \cite{Fujimoto2023TD7}, TD3-N \cite{ZhangTD3-N2024}, AC-Off-POC-TD3 \cite{Saglam2024onestepqlearning}, AC-TD3 \cite{jiang2024AC-TD3}, DivAC \cite{Yang2023DivAC}, EMDAC-TD3 \cite{shu2025episodic} for comparison. The specific features of these algorithms can be found in the respective references.

\begin{figure}[t]
  \centering
  \includegraphics[width = 0.48\textwidth]{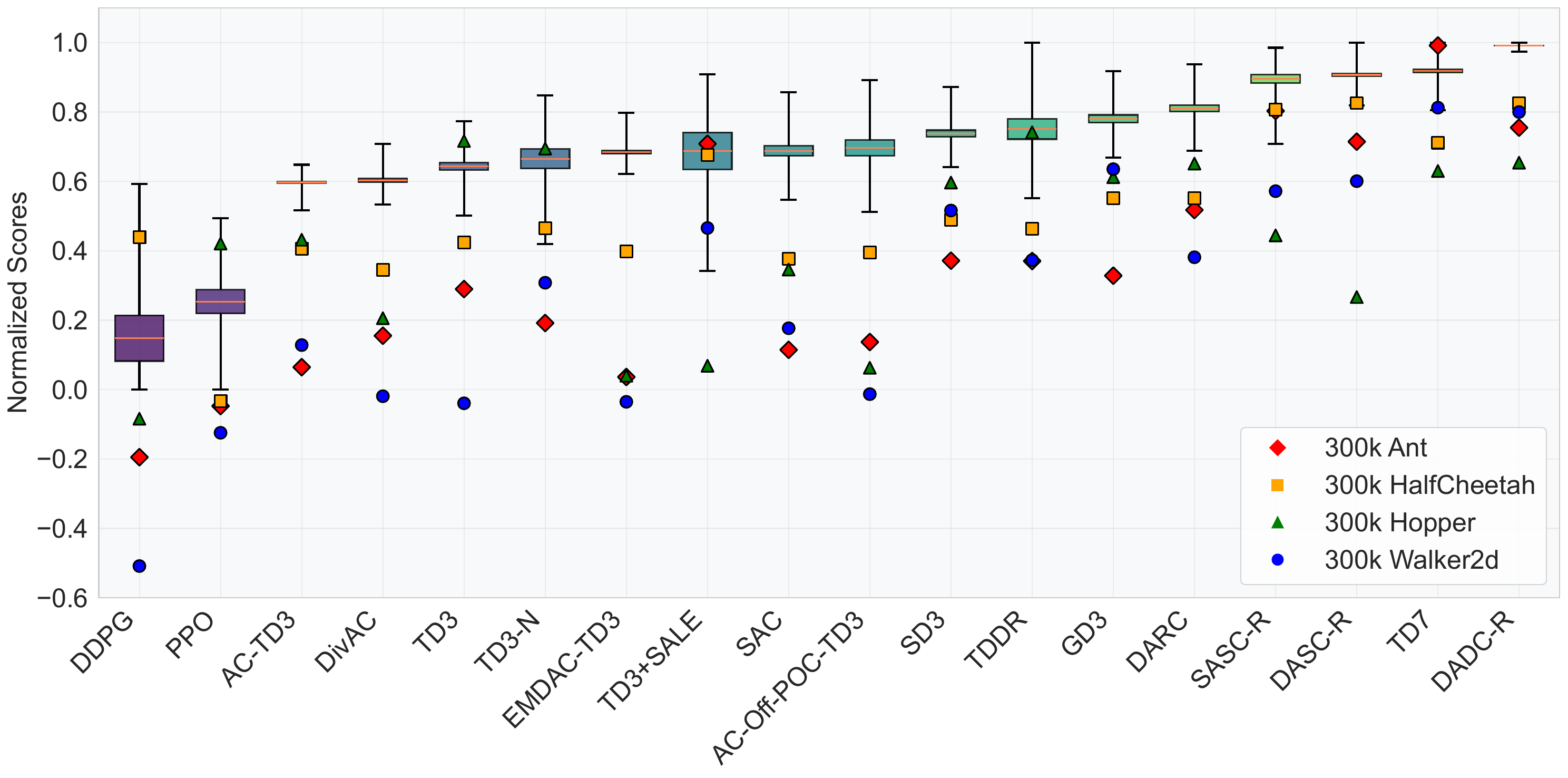}
\caption{Box plot illustrating the minimum, maximum, mean, and variance of the nine normalized average returns from the comparisons in Tables~\ref{tab:numerical_comparison_baseline_algorithm}, \ref{tab:numerical_comparison_algorithm}, and \ref{table.SOTA}. Red diamonds, orange squares, green triangles, and blue circles denote performance at 300k steps in Ant, HalfCheetah, Hopper, and Walker2d, respectively.}
  \label{fig:boxplot}
\end{figure}

Finally, all comparisons in Tables~\ref{tab:numerical_comparison_baseline_algorithm}, \ref{tab:numerical_comparison_algorithm}, and \ref{table.SOTA} across Ant, HalfCheetah, Hopper, and Walker2d are summarized in Fig.~\ref{fig:boxplot}, which presents box plots of the normalized average returns for each algorithm. For each environment, the returns of all algorithms are linearly normalized, with the highest return set to 1 and the lowest to 0. The orange line indicates the mean, while the upper and lower bounds of the box represent one variance above and below the mean, respectively. The horizontal lines at the top and bottom of the plot denote the maximum and minimum normalized scores, and the whiskers span the full range of observed values. A higher box position indicates superior overall performance. Colored markers denote performance at 300k steps in Ant, HalfCheetah, Hopper, and Walker2d, respectively, with higher marker positions reflecting greater sample efficiency.

As shown in Fig. \ref{fig:boxplot}, DADC-R achieves the highest mean among all evaluated algorithms and exhibits the shortest box plot, underscoring its superior performance. In comparison, DASC-R and SASC-R attain a mean lower only than that of TD7, with their relatively short box plots indicating consistent performance. DADC-R also demonstrates remarkable sampling efficiency, only performing below TD7 and TD3+SALE in Ant. From Table \ref{table.SOTA}, TD7 exhibits the largest standard deviation across Ant, HalfCheetah, and Walker2d. While DADC-R in HalfCheetah is slightly lower than that of DASC-R, it significantly surpasses other algorithms. In Ant, DADC-R is marginally below TD7, yet its smaller standard deviation reflects superior stability.

\begin{table*}[!t] 
  \caption{Average return and standard deviations for DADC-R, DASC-R, SASC-R, and SOTA algorithms}
  \centering
  \small
  \resizebox{\textwidth}{!}{
  \begin{tabular}{cccccccccccc}
    \toprule
    Algorithms & DADC-R & DASC-R & SASC-R & TD3+SALE & TD7 & TD3-N & AC-Off-POC-TD3 & AC-TD3 & DivAC & EMDAC-TD3 \\
    \midrule 
    Ant & 6958 $\pm$ 118 & 6452  $\pm$ 347 & 6590  $\pm$ 618 & 6391 $\pm$ 1121 & $\bm{6968 \pm 1905}$ & 3298 $\pm$ 162 & 3885 $\pm$ 287 & 3913 $\pm$ 792 & 4015 $\pm$ 614 & 4574 $\pm$ 656 \\
    HalfCheetah & 15570  $\pm$ 189 & $\bm{15672 \pm 152}$   & 15467  $\pm$ 292 & 14088 $\pm$ 920 & 14880 $\pm$ 2076 & 10090 $\pm$ 311 & 10030 $\pm$ 256 & 9934 $\pm$ 168 & 9190 $\pm$ 207 & 10764 $\pm$ 446 \\
    Hopper & $\bm{3581 \pm 124}$ & 3284  $\pm$ 411 & 3070  $\pm$ 474& 2368 $\pm$ 414 & 3257 $\pm$ 647 & 3338 $\pm$ 414 & 3424 $\pm$ 281 & 2926 $\pm$ 733 & 2925 $\pm$ 569 & 3242 $\pm$ 259 \\
    Walker2d & $\bm{5359\pm427}$ & 5000  $\pm$ 408 &  5189  $\pm$ 291  & 4087 $\pm$ 583 & 5114 $\pm$ 1406 & 4671 $\pm$ 476 & 4651 $\pm$ 149 & 4199 $\pm$ 820 & 4398 $\pm$ 371 & 4264 $\pm$ 454 \\
    BipedalWalker & $\bm{312 \pm 9}$ & $\bm{312 \pm 9}$ & $\bm{312 \pm 9}$  & 305 $\pm$  16 & 304 $\pm$ 48 & 296 $\pm$ 32 & 307 $\pm$ 18 & 298 $\pm$ 33 & 292 $\pm$ 38 & 308 $\pm$  20 \\
    InvertedDoublePendulum & 9345 $\pm$ 17 & 9345 $\pm$ 17 & 9345 $\pm$ 17  & 7634 $\pm$ 3012 & $\bm{9357 \pm 4}$ & 9154 $\pm$ 534 & 9327 $\pm$ 12 & 8971 $\pm$ 1037 & 8972 $\pm$ 991 & 9322 $\pm$  29 \\
   \bottomrule
  \end{tabular}
  }
  \label{table.SOTA}
\end{table*}

\section{Conclusion}
\label{sec:Conclusion}

In this paper, we introduce a novel framework that enhances TDDR by addressing two fundamental challenges in deep reinforcement learning: flexible estimation bias control and high-quality representation learning. For bias control, we propose a set of three distinct convex combination strategies, governed by a single hyperparameter, that synergistically integrate pessimistic and optimistic estimates. This mechanism provides tunable control over the full spectrum of estimation biases. For performance enhancement, we integrated a predictive representation learning module that learns high-quality state and action representations from low-level inputs, serving as inputs for the actor and critic networks. The representation-enhanced variants significantly outperform strong baselines and several SOTA methods. Importantly, our analysis of $\upsilon$ reveals that reducing estimation bias does not always lead to improved value estimation, as both overestimation and underestimation can be exploited differently depending on the environment.

For future research, it would be interesting to explore adaptive scheduling of $\upsilon$.

\bibliographystyle{IEEEtran}
\bibliography{IEEEabrv,IEEETrans}

\end{document}